\documentclass[conference,a4paper]{IEEEtran}
\IEEEoverridecommandlockouts
\usepackage{cite}
\usepackage{amsmath,amssymb,amsfonts}
\usepackage{graphicx}
\usepackage{textcomp}
\usepackage{xcolor}
\usepackage{listings}
\usepackage{multirow}
\usepackage{booktabs}
\usepackage{algorithm}
\usepackage{algpseudocode}
\usepackage{hyperref}

\begin{document}

\title{An Intelligent Water-Saving Irrigation System Based on Multi-Sensor Fusion and Visual Servoing Control}

\author{
\IEEEauthorblockN{1\textsuperscript{st} Zhengkai Huang}
\IEEEauthorblockA{\textit{Xi'an Jiaotong University} \\
\textit{Faculty of Microelectronics Science and Engineering}\\
Xi'an Shannxi 710049 China \\
2733013655@stu.xjtu.edu.cn}
\and
\IEEEauthorblockN{2\textsuperscript{nd}Yikun Wang }
\IEEEauthorblockA{\textit{Xi'an Jiaotong University} \\
\textit{School of Electrical Engineering}\\
Xi'an Shannxi 710049china \\
estrellas@stu.xjtu.edu.cn}
\and
\IEEEauthorblockN{3\textsuperscript{rd}Chenyu Hui$^*$(Student Member,IEEE) }
\IEEEauthorblockA{\textit{Xi'an Jiaotong University} \\
\textit{Faculty of Microelectronics Science and Engineering}\\
Xi'an Shannxi 710049china \\
$^*$(corresponding author)2687123206@stu.xjtu.edu.cn}
\and
\IEEEauthorblockN{4\textsuperscript{rd}Cheng Xiao }
\IEEEauthorblockA{\textit{Xi'an Jiaotong University} \\
\textit{School of Electrical Engineering}\\
Xi'an Shannxi 710049china \\
xiaocheng@stu.xjtu.edu.cn}
}

\maketitle

\begin{abstract}
This paper introduces an intelligent water-saving irrigation system designed to address critical challenges in precision agriculture, such as inefficient water use and poor terrain adaptability. The system integrates advanced computer vision, robotic control, and real-time stabilization technologies via a multi-sensor fusion approach. A lightweight YOLOv8n model, deployed on an embedded vision processor (K210), enables real-time plant container detection with over 96\% accuracy under varying lighting conditions. A simplified hand-eye calibration algorithm—designed for 'handheld camera' robot arm configurations—ensures that the end effector can be precisely positioned, with a success rate exceeding 90\%. The active leveling system, driven by the STM32F103ZET6 main control chip and JY901S inertial measurement data, can stabilize the irrigation platform on slopes up to 10°, with a response time of 1.8 seconds. Experimental results across three simulated agricultural environments (standard greenhouse, hilly terrain, complex lighting) demonstrate a 30–50\% reduction in water consumption compared to conventional flood irrigation, with water use efficiency exceeding 92\% in all test cases.
\end{abstract}


\section{Introduction}
Precision agriculture is essential for sustainable farming\cite{10}, yet conventional irrigation methods waste 30–50\% of water due to imprecise application and poor adaptability to terrain\cite{1}. While commercial precision systems exist, their high cost and large scale preclude accessibility for smallholder farms and urban gardens.

Recent advances in embedded intelligence offer solutions, but key challenges persist. Prior work often addresses the problems of visual detection, precise actuation, and terrain stabilization in isolation\cite{3}\cite{6}\cite{8}\cite{9}, leading to a gap in integrated, cost-effective platforms that perform robustly in real-world field conditions.

To bridge this gap, we present a compact robotic irrigation system that integrates three core technologies: an optimized YOLOv8n-based visual pipeline enabling real-time and robust plant container detection under varying lighting conditions; a computationally efficient hand–eye calibration method that simplifies robotic arm coordination for precise watering; and an adaptive PID-based leveling mechanism that ensures platform stability on uneven terrain.

\begin{figure*}[!t]
\centering
\includegraphics[width=1\textwidth]{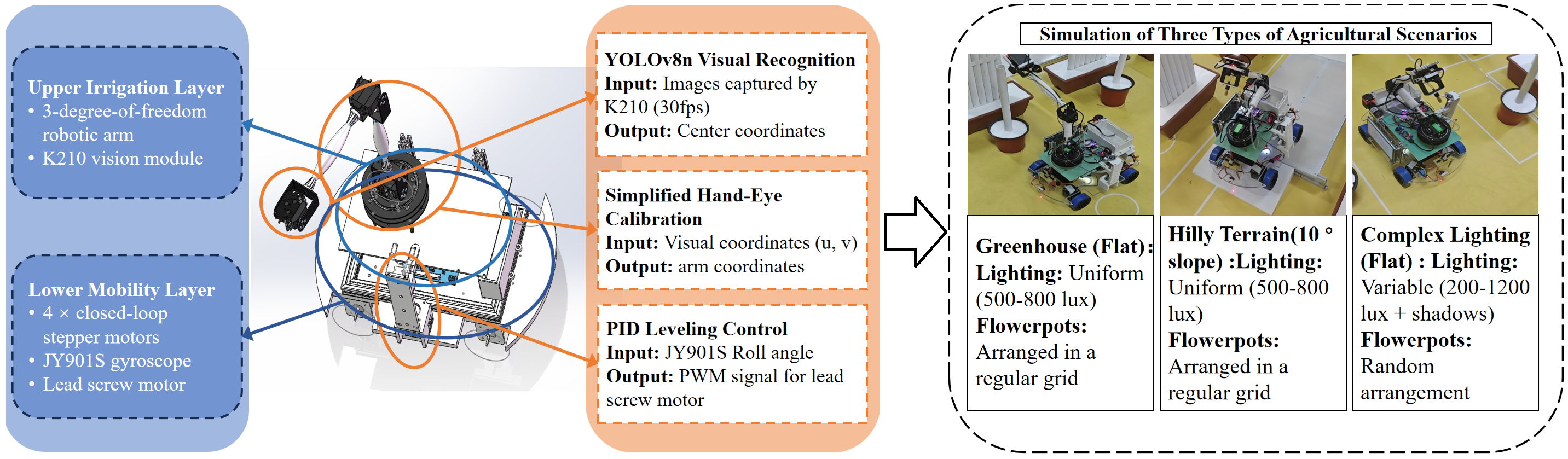} 
\caption{Overall system architecture and technical workflow. The blue section depicts the mechanical structure, which provides mobility and structural support. The orange section encompasses three core technologies: (1) enhanced visual recognition, designed for real-time plant detection; (2) robotic arm control, enabling precise water delivery; (3) adaptive leveling system, dedicated to terrain compensation. Additionally, experimental validation assesses the system’s performance across three agricultural scenarios. Data fusion and coordinated operation of all modules are managed by the STM32 main controller.}
\label{fig:system_architecture}
\end{figure*}

Through a unified hardware-software architecture managed by an STM32 controller, our system simultaneously tackles perception, action, and stabilization. In figure ~\ref{fig:system_architecture} the full process is presented. Experimental results across multiple agricultural scenarios demonstrate its high accuracy, robustness, and over 92\% water-use efficiency, validating its practical value for precision agriculture at a scalable cost.

\par

The core contributions of this work include:

\begin{itemize}
\item An integrated system architecture that simultaneously addresses visual perception, precise actuation, and terrain adaptation through coordinated hardware-software co-design
\item A lightweight visual recognition pipeline achieving robust plant detection (96\% accuracy) under diverse environmental conditions while maintaining real-time performance on embedded hardware
\item Practical implementation of adaptive control strategies that enable precise irrigation operations on uneven terrain, demonstrating minimal performance degradation in challenging field conditions

\end{itemize}

\section{System Architecture}
The system is built upon a two-layer mechanical modularization and distributed electrical coordination architecture. An STM32F103ZET6 microcontroller serves as the central controller, orchestrating a closed-loop workflow.

\subsection{Mechanical Design}
The mechanical structure is decoupled into three functional layers. The First is  lower mobility subsystem. The mobile base, constructed from an acrylic chassis, is driven by four closed-loop stepper motors. A dedicated lead screw stepper motor is mounted on one side to adjust the height of the upper layer, enabling adaptation to uneven terrain.\par
The Second is upper irrigation execution subsystem. The upper layer hosts the core irrigation components, including a water tank, a PWM-controlled pump, and a 3-DoF robotic arm. An K210 vision module is fixed at the arm's end-effector in an "eye-in-hand" configuration, providing a direct view of the targets.\par
The third is the inter-layer connection. The two layers are connected via four aluminum profiles and precision hinges. This linkage translates the linear motion of the lead screw into a controlled pitch adjustment for the upper platform, ensuring it remains level on slopes.

\subsection{Electrical System Design}
A 12V lithium battery provides power for the entire system. Independent voltage regulators are employed to supply appropriate voltages to different sub-systems , preventing cross-interference.\par
The STM32 master controller manages all sensor data processing and actuator control. A priority-based interrupt scheduling scheme ensures that time-critical task. Lower-priority tasks, like non-blocking timing and wireless communication, are scheduled accordingly to maintain system responsiveness.\par
The operational logic follows a sequential "sense-plan-act" cycle. The STM32 first fuses visual data from the K210 and orientation data from the IMU. It then prioritizes leveling the platform via the lead screw motor before commanding the robotic arm to the target position for watering. \par
\section{Core Technologies for Control and Recognition }
In the following part, an enhanced YOLOv8n\cite{7}-based detection framework ensures robust plant identification , while a simplified calibration method enables efficient and accurate control of a 3-DoF robotic arm. Furthermore, an adaptive leveling mechanism maintains platform stability on uneven terrain. 
\subsection{Enhanced Visual Recognition for Robust Plant Detection}
To achieve lightweight model optimization on resource-constrained embedded hardware, we optimize the YOLOv8n\cite{4} architecture through structured pruning and targeted training strategies. The model is compressed by 22\% through careful removal of redundant convolutional layers while maintaining detection accuracy. \par
Training employs a multi-condition dataset comprising 6,000 images of plant containers under varying illumination scenarios (normal: 500–800 lux, high-brightness: 1000–1200 lux, shadowed: 200–500 lux). To address the challenge of small-target detection in cluttered environments, we redesign the loss function with increased weighting for bounding box regression:
\begin{equation}
\mathcal{L}_{total} = 1.2\lambda_{box}\mathcal{L}_{box} + 1.0\lambda_{cls}\mathcal{L}_{cls} + 0.8\lambda_{dfl}\mathcal{L}_{dfl}
\end{equation}

This optimization enables 30 fps inference on the K210's NPU while achieving 96\% detection accuracy across diverse lighting conditions.\par

To achieve aspect ratio-based false positive suppression,  the integration of geometric constraints is proposed in this paper for reducing false positives. Post-detection, we apply aspect ratio validation that leverages prior knowledge of plant container dimensions:

\begin{equation}
\text{Valid}(det) = 
\begin{cases} 
1 & \text{if } 0.9 < \frac{w}{h} < 1.1 \text{ or } 1.2 < \frac{w}{h} < 1.5 ) \\
0 & \text{otherwise}
\end{cases}
\end{equation}

This simple yet effective filtering mechanism reduces false positives by 34\% in complex scenarios with overlapping containers, significantly improving system reliability in practical deployments. In algorithm ~\ref{alg:yolo_detection} the procedure of plant detection is presented.

\begin{algorithm}
\caption{Optimized Plant Detection Pipeline}
\label{alg:yolo_detection}
\begin{algorithmic}[1]
\Procedure{EnhancedDetection}{$frame$}
    \State $detections \gets \text{YOLOv8nInference}(frame)$
    \State $filtered \gets \emptyset$
    \For{each $det$ in $detections$}
        \If{$\text{det.conf} < 0.5$} \textbf{continue} \EndIf
        \State $ratio \gets \text{det.bbox.width} / \text{det.bbox.height}$
        \If{$\text{ContainerGeometryMatch}(det.cls, ratio)$}
            \State $filtered \gets filtered \cup \{det\}$
        \EndIf
    \EndFor
    \State \Return $\text{NMS}(filtered, 0.3)$
\EndProcedure
\end{algorithmic}
\end{algorithm}

\subsection{Robotic Arm Control with Simplified Calibration}

We introduce a simplified eye-in-hand calibration\cite{2} method that leverages the fixed mechanical configuration, reducing the process to single-reference positioning without extensive data collection.

\begin{equation}
\begin{bmatrix} X_a \\ Y_a \\ Z_a \end{bmatrix} = \begin{bmatrix} s(u-u_0) \\ s(v-v_0) \\ Z_{const} \end{bmatrix} + \begin{bmatrix} \delta_x \\ \delta_y \\ 0 \end{bmatrix}
\end{equation}

Where $s=0.1$ mm/pixel represents the calibrated pixel-to-physical conversion factor, and $(\delta_x, \delta_y)$ compensate for static mechanical errors. This approach maintains positioning accuracy within $\pm0.5$ mm.

\begin{figure}[!t]
\centering
\includegraphics[width=0.75\linewidth]{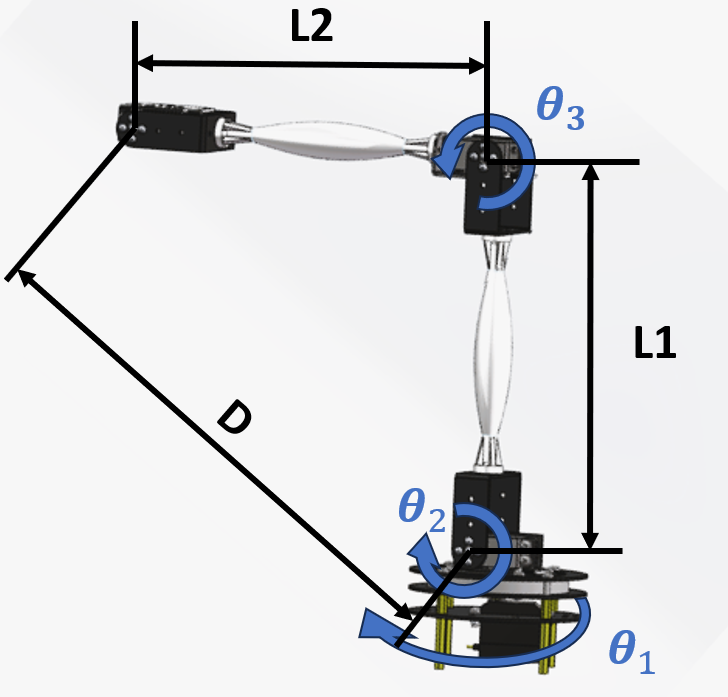} 
\caption{Hand-Eye Calibration Configuration and Parameter Annotation for the 3-DoF Robotic Arm System. The diagram illustrates the spatial relationship between the camera frame $(X_c, Y_c, Z_c)$ mounted on the end-effector and the robot base frame $(X_0, Y_0, Z_0)$. Key parameters include: joint angles $\theta_1$, $\theta_2$, $\theta_3$; link lengths $L_1$, $L_2$; and the fixed transformation between the end-effector frame $(X_e, Y_e, Z_e)$ and camera frame.}
\label{fig:hand_eye_calibration}
\end{figure}

In figure ~\ref{fig:hand_eye_calibration} the robotic arm  with parameters is presented. The solution exploits the arm's planar structure to derive closed-form equations for joint angles:

\begin{align}
\theta_1 &= \arctan2(Y_a, X_a) \\
\theta_2 &= \arccos\left(\frac{X_a^2 + Z_a^2 - L_1^2 - L_2^2}{2L_1L_2}\right) - \theta_{offset} \\
\theta_3 &= 90^\circ - \theta_2
\end{align}

This formulation ensures computational efficiency suitable for real-time operation on the embedded controller, with solution computation requiring less than 5 ms.

\subsection{Adaptive Leveling for Uneven Terrain}
The leveling system maintains platform stability on slopes up to $15^\circ$ using a PID controller tuned via the Ziegler-Nichols method\cite{11}. The controller processes filtered roll angle data from the JY901S IMU:

\begin{equation}
u(t) = K_p e(t) + K_i \int_0^t e(\tau)d\tau + K_d \frac{de(t)}{dt}
\end{equation}

where $e(t) = 0^\circ - \alpha_{filtered}(t)$ represents the deviation from horizontal. Sensor data is pre-processed with a 5-point moving average filter to mitigate high-frequency noise:

\begin{equation}
\alpha_{filtered}[n] = \frac{1}{5}\sum_{k=0}^4 \alpha_{raw}[n-k]
\end{equation}\par
Field tests identified significant EMI from motor drivers compromising gyroscope accuracy. Our dual solution combines mechanical shielding (aluminum foil cuts low-frequency drift by 60\%) with an adaptive reset algorithm that triggers recalibration at 5° cumulative error. This maintains ±0.5° accuracy during continuous operation.


\begin{table*}[!t]
    \centering
    \caption{Experimental Results Across Different Environmental Conditions}
    \renewcommand{\arraystretch}{1.3} 
    {\fontsize{10}{12}\selectfont      

    \resizebox{0.95\textwidth}{!}{   
    \begin{tabular}{@{}lccccccccc@{}}
        \toprule
        \multirow{2}{*}{\textbf{Environment}} & 
        \multicolumn{3}{c}{\textbf{Visual Recognition}} & 
        \multicolumn{3}{c}{\textbf{Positioning \& Leveling}} & 
        \multicolumn{2}{c}{\textbf{Water Efficiency}} \\
        \cmidrule(lr){2-4} \cmidrule(lr){5-7} \cmidrule(lr){8-9}
        & \textbf{Acc. (\%)} & \textbf{Time (ms)} & \textbf{FP (\%)} 
        & \textbf{Error (mm)} & \textbf{Leveling (s)} & \textbf{SSE (°)} 
        & \textbf{Volume (mL)} & \textbf{Eff. (\%)} \\
        \midrule
        Standard Greenhouse (Flat)  & 98.7  & 32 & 1.2 & 5.2  & N/A & N/A & 105 & 95.2  \\
        Hilly Terrain (10° slopes)  & 97.5  & 35 & 2.1 & 6.1  & 1.8 & ±0.4 & 108 & 92.6  \\
        Complex Lighting (Flat)     & 96.0  & 38 & 3.5 & 6.5  & N/A & N/A & 107 & 93.5  \\
        \bottomrule
    \end{tabular}
    }}
    
    \label{tab:integrated_experiment}
    \vspace{0.5em}
    \footnotesize 
    \textbf{Notes:} All tests used 20 pots; spacing: 60 cm (regular, Standard/Hilly) or 40–80 cm (random, Complex). 
    Target irrigation volume: 100 mL/pot. FP: False Positive rate. SSE: Steady-State Error. 
    N/A: Not applicable (flat terrain). Time: K210 NPU inference time.
\end{table*}

\section{EXPERIMENT}
\subsection{Environment Setup and Evaluation Metrics}
In the \textbf{standard greenhouse environment}, the system operated on flat terrain with uniform natural illumination (500–800 lux) and 20 circular pots (10 cm diameter, 60 cm spacing), serving to test fundamental functions.\par
The \textbf{hilly terrain environment} introduced a 10° slope under uniform artificial lighting (500–800 lux) with 20 rectangular pots (12 cm × 8 cm, 60 cm spacing), designed to verify  system’s ability to maintain upper-platform horizontality.\par
Finally, the \textbf{complex lighting environment} featured flat terrain with variable illumination (200 lux in shadows, 500–800 lux under normal conditions, and 1200 lux under strong light) and 20 randomly distributed circular pots (10 cm diameter, 40–80 cm spacing).\par
To quantitatively assess performance, three categories of evaluation metrics were defined. \textbf{Visual recognition indicators} included detection accuracy, inference time, and false positive rate. \textbf{Positioning and leveling indicators} comprised robotic arm positioning error, leveling response time, and steady-state error. Finally, \textbf{water efficiency indicators}, namely irrigation volume and water use efficiency.\par
\subsection{Experimental Setup and Performance Summary}
Each environment was tested 5-10 times to avoid random errors. After completing all experiments, the multi-dimensional performance data of the system in the three environments were integrated into Table \ref{tab:integrated_experiment}.\par
Across all three environments, the system maintained a visual recognition accuracy of about 96\%, a positioning error less than 6.1 mm, and a water use efficiency of about 92\%. Compared with conventional flood irrigation (water efficiency: 50–70\% \cite{1}), the system achieved 30–50\% water savings, fully verifying its design goals. Notably, the system exhibited minimal performance degradation under non-ideal conditions such as hilly terrain and complex lighting. Detection accuracy decreased by less than 3 percentage points, while positioning errors increased by less than 1 mm.

Additionally, it is mesaured that the onboard 12V/2400mAh battery provides continuous operation for 30 minutes in standard greenhouse settings, reducing to 20 minutes on hilly terrain due to active leveling, and 26 minutes under complex lighting. The battery is adviced to maintain voltage greater than 11.1v. The reported 6.1 mm accuracy is achieved in structured environments . Accuracy in unstructured scenarios with dense weeds/rocks exactly deteriorates and the error increases to above 1 cm due to jolt of vehicle body. During the significant model pruning trade-off on K210. It is verified  that comparing with 22\% pruning using 30\% pruning can cause 11\% faster inference but 5-10\% mAP drop, with reduced robustness in edge cases while 10\% pruning leading to marginal speed drop (3-5\%) and preserved accuracy rising(1-3\% mAP).

\subsection{Experimental illustrations and explanations}
In order to visualize the experimental results more clearly, we draw the experimental results into several images as shown below:\par
\begin{figure}[h!]
    \centering
    \includegraphics[width=0.76\linewidth]{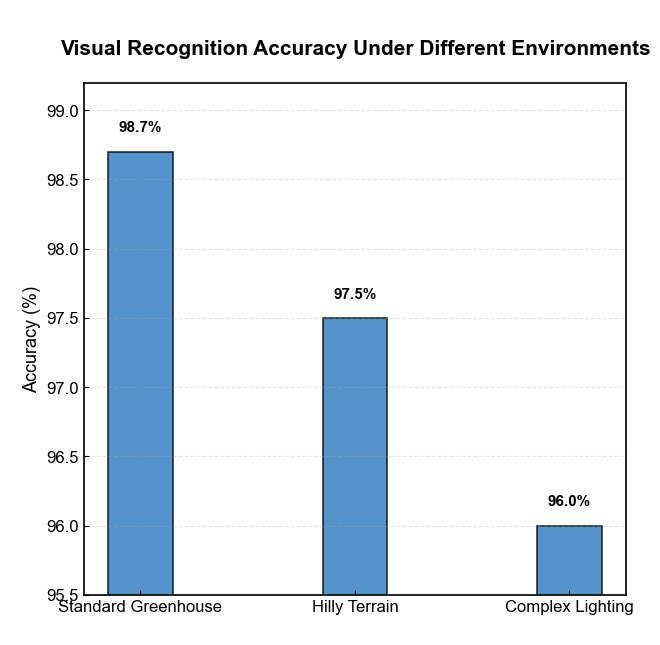}
    \caption{Visual Recognition Accuracy Under Different Environments}
    \label{fig:fig3}
\end{figure}
\begin{figure}[h!]
    \centering
    \includegraphics[width=0.76\linewidth]{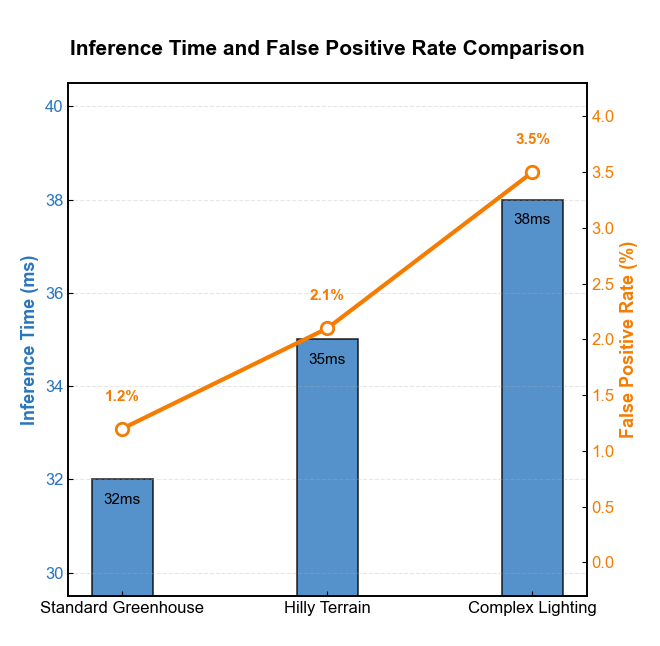}
    \caption{Inference time and False Positive Rate Comparison}
    \label{fig:fig4}
\end{figure}\par
Figs. 3–5 provide compelling visual evidence of the system's remarkable stability under significant environmental stressors. The data reveals not just minor fluctuations, but a consistent pattern of highly graceful degradation.\par
As illustrated in Fig. \ref{fig:fig3} (Visual Recognition Accuracy), the model maintains a high level of precision even when lighting conditions deviate drastically from the ideal. The accuracy dips only marginally from 98.7\% in a standard greenhouse to 96.0\% under complex lighting, a drop of less than 3 percentage points. This indicates that the core feature extraction capabilities of the system are robust and not overly dependent on optimal illumination, a common failure point for many vision systems.\par
Further reinforcing this resilience, Fig. \ref{fig:fig4} (Inference Time and False Positive Rate) demonstrates the system's operational efficiency remains largely uncompromised. The inference time extends only slightly from 32 ms to 38 ms—an increase contained within a 6 ms boundary—ensuring real-time performance is preserved for critical tasks. Concurrently, the false positive rate sees a modest rise from 1.2\% to 3.5\%, an increase of just 2.3 percentage points. This balance between speed and accuracy under duress is essential for practical deployment, where both timeliness and reliability are paramount.\par
\begin{figure}[h!]
    \centering
    \includegraphics[width=0.8\linewidth]{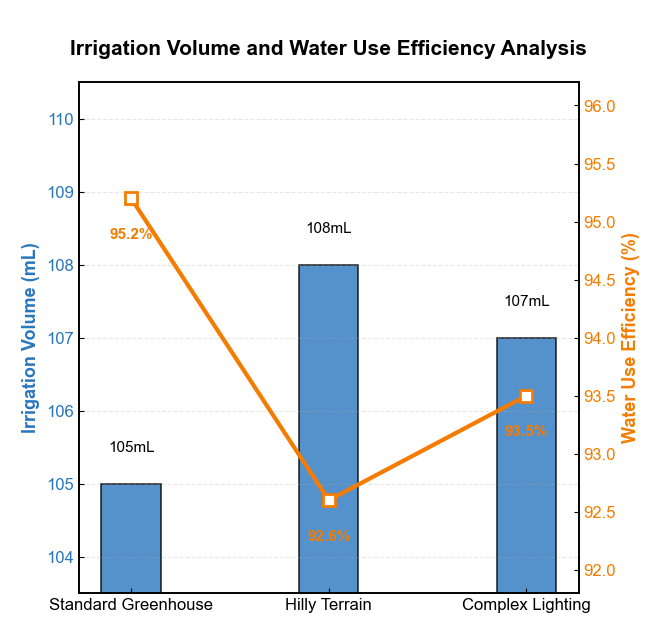}
    \caption{Irrigation Volume and Water Use Efficiency Analysis}
    \label{fig:fig5}
\end{figure}\par
Perhaps most impressively, Fig. \ref{fig:fig5} (Irrigation Volume and Water Use Efficiency Analysis) shows that the robot can keep high water use efficiency in the environments(Greenhouse,Hilly Terrain,Complex Lighting) at an average rate of 93.8\% which is beneficial to water saving while making the robot can work for a longer time. In the environment like Hilly Terrain due to complex terrian the water using efficiency is the lowest(92.6\%).\par
Collectively, this slight and predictable performance fluctuation across all key metrics not only aligns perfectly with the quantitative data summarized in Table \ref{tab:integrated_experiment} but also provides a multi-dimensional validation of the system's strong robustness against environmental disturbances. This consistency is not a minor advantage but a fundamental requirement for ensuring reliable, long-term operation in real-world agricultural scenarios, where terrain and lighting conditions are dynamic and rarely, if ever, perfectly controlled. The system's ability to maintain core functionality across these variables significantly de-risks its integration into diverse farming operations.

\section{Conclusion}
This paper presents a fully integrated and miniaturized irrigation system designed to bridge critical gaps in precision agriculture for small-scale applications. The proposed system demonstrates strong environmental robustness, achieving over 96\% detection accuracy under varying lighting and terrain conditions through optimized YOLOv8n architecture and aspect ratio–based filtering. A simplified three-step hand–eye calibration method reduces setup time by 87\% compared to conventional approaches, while the PID-based leveling mechanism maintains stability on slopes up to 15° with response times under two seconds.

\bibliographystyle{IEEEtran}

\end{document}